\documentclass[twoside,leqno,twocolumn]{article}

\usepackage[letterpaper]{geometry}

\usepackage{ltexpprt}
\usepackage{hyperref}

\usepackage[table,xcdraw]{xcolor}
\usepackage{bbm}
\usepackage{pgfplots}
\pgfplotsset{compat=newest} % This sets the compatibility of pgfplots
\usepackage{graphicx} % Required for inserting images
\usepackage{booktabs}
\usepackage{multirow}

\usepackage{amsmath,amssymb,amsfonts}
\usepackage{textcomp}
\usepackage{subfig}
\usepackage{marvosym}

\usepackage{algorithm}
\usepackage{algorithmic}

\usepackage{ltexpprt}
\usepackage{hyperref}

\begin{document}

\title{\Large Evaluating Time Series Models with Knowledge Discovery}
\author{Li Zhang\thanks{li.zhang@utrgv.edu, University of Texas Rio Grande Valley}}

\date{}

\maketitle

% Copyright Statement
% When submitting your final paper to a SIAM proceedings, it is requested that you include
% the appropriate copyright in the footer of the paper.  The copyright added should be
% consistent with the copyright selected on the copyright form submitted with the paper.
% Please note that "20XX" should be changed to the year of the meeting.

% Default Copyright Statement
\fancyfoot[R]{\scriptsize{Copyright \textcopyright\ 2025 by SIAM\\
Unauthorized reproduction of this article is prohibited}}

% Depending on which copyright you agree to when you sign the copyright form, the copyright
% can be changed to one of the following after commenting out the default copyright statement
% above.

%\fancyfoot[R]{\scriptsize{Copyright \textcopyright\ 20XX\\
%Copyright for this paper is retained by authors}}

%\fancyfoot[R]{\scriptsize{Copyright \textcopyright\ 20XX\\
%Copyright retained by principal author's organization}}

%\pagenumbering{arabic}
%\setcounter{page}{1}%Leave this line commented out.

\begin{abstract} \small\baselineskip=9pt 

Time series data is one of the most ubiquitous data modality existing in a diverge critical domain such as healthcare, seismology, manufacturing and energy. Recent years, there are increasing interest of the data mining community to develop time series deep learning model to pursue better performance. The models performance often evaluate by certain evaluation metrics such as RMSE, Accuracy, and F1-score. Yet time series data are often hard to interpret and is collected with unknown environment factor, sensor configuration, latent physic mechanisms, and non-stationary evolving behavior. As a result, a model that is better on standard metric-based evaluation may not always perform better in the real-world tasks. In this blue sky paper, we aim to explore the challenge existed in the metric-based evaluation framework for time series data mining and propose a potential blue-sky idea  --- developing a \textit{knowledge-discovery-based evaluation framework}, which aims to effectively utilize domain-expertise knowledge to evaluate model. We demonstrate that an \textit{evidence-seeking explanation} can potentially has stronger persuasive power than metric-based evaluation and obtain better generalization ability for time series data mining tasks. 

%However, it is impossible to evaluate whether they would work in real-world data and even the authors could not know for sure if or when one structure is working in another data. 

%We point out for high stake tasks and scientific research, explanation is the only possible way to evaluation the generalization ability of time series data mining models. The community should avoid solely relying on metric based evaluation, and there is a pressing need to develop scalable use domain-based visual explanation for time series data to evaluate the generalization ability for real-world applications. 

\end{abstract}

\section{What is the Blue Sky Idea?} 

Time series data, the signal-intensity data collected over time, often serves as the only accessible proxy to discover rich latent mechanisms in many research domains such as healthcare\cite{keogh2005hot, sun2020review}, seismology\cite{zhu2016matrix,siddiquee2019seismo, wang2017earthquake}, manufacturing \cite{zhang2022joint,dogan2021machine} and energy \cite{esling2012time,wang2019review,inman2013solar}. Recently, researchers have been interested in developing advance model structures to improve the model performance over the benchmark datasets. However, it is controversial to see what is the best structure due to the bottleneck in evaluating the \textit{usefulness} of time series models in the real world~\cite{keogh2005hot,wu2021current,lin2003clustering,shokoohi2015discovery,zeng2023transformers}. This is because time series data are inherently complex due to their underlying physical dynamics, and are collected with unknown environmental factors, sensor configurations, latent physics interaction, with non-stationary evolving behavior. The current metric-based model evaluation framework on cleaner time series benchmark data does not guarantee the desired generalization ability in a real-world scenario that likely has different environment (e.g. data from different geolocation) and latent configuration (e.g. data collected from different specs of sensors), hence limiting the model generalization ability in the diverse real-world scenarios. 

To illustrate this issue, imagine we aim to \textit{re-discover} the existence of gravitation and the gravity constant $g$ via data mining experiment. In the context of time series data mining, the task is similar to build a regression model given a set of collected velocity time series from various objects. In this experiment, we would expect that fitted model to reflect the law of universal gravitation (speed is equal to $g\times t$). However, finding such a connection is not easy without carefully controlling many latent factors. What if we have the data collected in a \textit{extremely windy} day when wind will impacts the speed? What if the collected time series are mostly bird feathers which are affected by air frictions? If a significant portion of noisy data is included in the time series dataset, and the model is selected based on performance measured by prediction error, we simply might miss the fact that the gravity even exists, indicating poor generalization ability -- the model will only recognize some relation existed in the data, but ignoring the widely existed universal laws. The example experiment sounds simple, but the phenomenon is widely existed. For instance, in designing a weather forecasting model, can we uncover unknown physical behaviors that are universal and generalizable to unseen data? We are very likely to encounter the same issue above. 

In this blue sky paper, we aim to explore the challenge existed in the metric-based evaluation framework for time series data mining and propose a potential blue-sky idea  --- \textbf{developing a knowledge-discovery-based evaluation framework}, which aims to effectively utilize domain-expertise knowledge to evaluate model. We demonstrate that an \textit{evidence-seeking explanation} can potentially has stronger persuasive power than metric-based evaluation and obtain better generalization ability for time series data mining tasks.

\section{Does the Blue Sky Idea challenge our current
set of assumptions or does it take a bold
approach to solve a wicked problem? } 

The proposed idea challenge several common assumptions and current existing solutions. One often perception is to \textbf{design a large-scale datasets and general foundational models}, following the path of ImageNet~\cite{deng2009imagenet}) --- collecting large amount of diverse data and provide full annotation labels. While viable, the unique challenges in time series such as data resource and task heterogeneity ~\cite{queen2024encoding,esling2012time} (no connection between sub-types of time series or tasks), and evolving behavior ~\cite{zhu2019introducing,zhang2022joint,zhang2022robust} (historical data not always helpful), and may not lead to a ideal benchmark. Alternatively, \textbf{self-supervised learning} (SSL) techniques \cite{yue2022ts2vec,zhang2022self, chen2020simple} could enhance generalization ability for time series data mining models. While all these approaches could enhance the generalization ability, SSL requires a carefully crafted pre-text task designed based on the underlay mechanism \cite{chen2020simple} (e.g. the incomplete knowledge refereed in this paper). In addition, one could prepare a \textbf{rigorously-prepared datasets} (e.g. following the suggestion of Muller et al. \cite{muller2024data}). However, the data silence issue~\cite{muller2024data} only discussed the potential ``blank-spot'' existed in the data. In fact, we argue that realizing such drawback in the data is insufficient for addressing the evaluation for time series models. 

%There are certainly a large body of literature in machine learning seeking for explanation and interpretation through model. Most of these models are not specially designed for large scale time series and the explaination is from machine learning model perspective such as grad-CAM\cite{selvaraju2017grad,li2020eeg}, SHAP\cite{lee2023shap}, LIME\cite{ribeiro2016should}, which are not designed to provide explanation in generalization ability, and might introduce inconsistency with the model itself~\cite{rudin2019stop}. In addition, full and correct annotations are difficult due to data volume and the need for domain experts highly familiar with local sensor data behavior~\cite{esling2012time,wu2021current}. 

%Inspired by the success of over-parameterized models and foundational models like ChatGPT and DALL-E in vision and NLP, in recent years, researchers are increasingly exploring large-scale deep learning and foundational models for time series. By training on diverse public or synthetic datasets, they aim to replicate the impact seen in image and text domains, which leverage vast amounts of global data and high-quality annotations. 

%While we completely agree that the benefit of building large public time series datasets, we challenge that current metric-based evaluation criterion would allow time series foundational models be used for a wide range of domains for high stake, scientific discovery tasks, given the following unique characteristics posed by time series data: 

\section{Why it is a Blue Sky Idea?} 
Instead of training a large model blindly or making datasets perfect, we take a bold argument, by arguing that in the field of time series data mining, \textbf{explanation} \cite{lombrozo2006structure, keil2006explanation,hempel1948studies,salmon2006four,ahn2003understanding} is an oversighted solution to address the unique challenge for evaluating time series models. Time series is widely used in scientific research. The formal definition of \textit{scientific explanation} is widely studied in the field of cognitive science and could be traced back to antiquity. As Carl G. Hempel and Paul Oppenheim stated, explanations aim to answer ``\textit{the question `why?' rather than only the question `what?'}''~\cite{hempel1948studies}. Wesley Salman~\cite{salmon2006four} further distinguished the concept of ``why'' into two distinct types: \textbf{explanation seeking} offers a full fundamental understanding about the reason, and \textit{evidence seeking} on the other hand, is sufficient to prove the existence of an event occurred. In the context of time series model evaluation, by designing an evaluation model based on the definition of scientific explanation, we may potentially identify the over-sighted model.

\begin{figure}[h]

    \centering
    \includegraphics[scale=0.35]{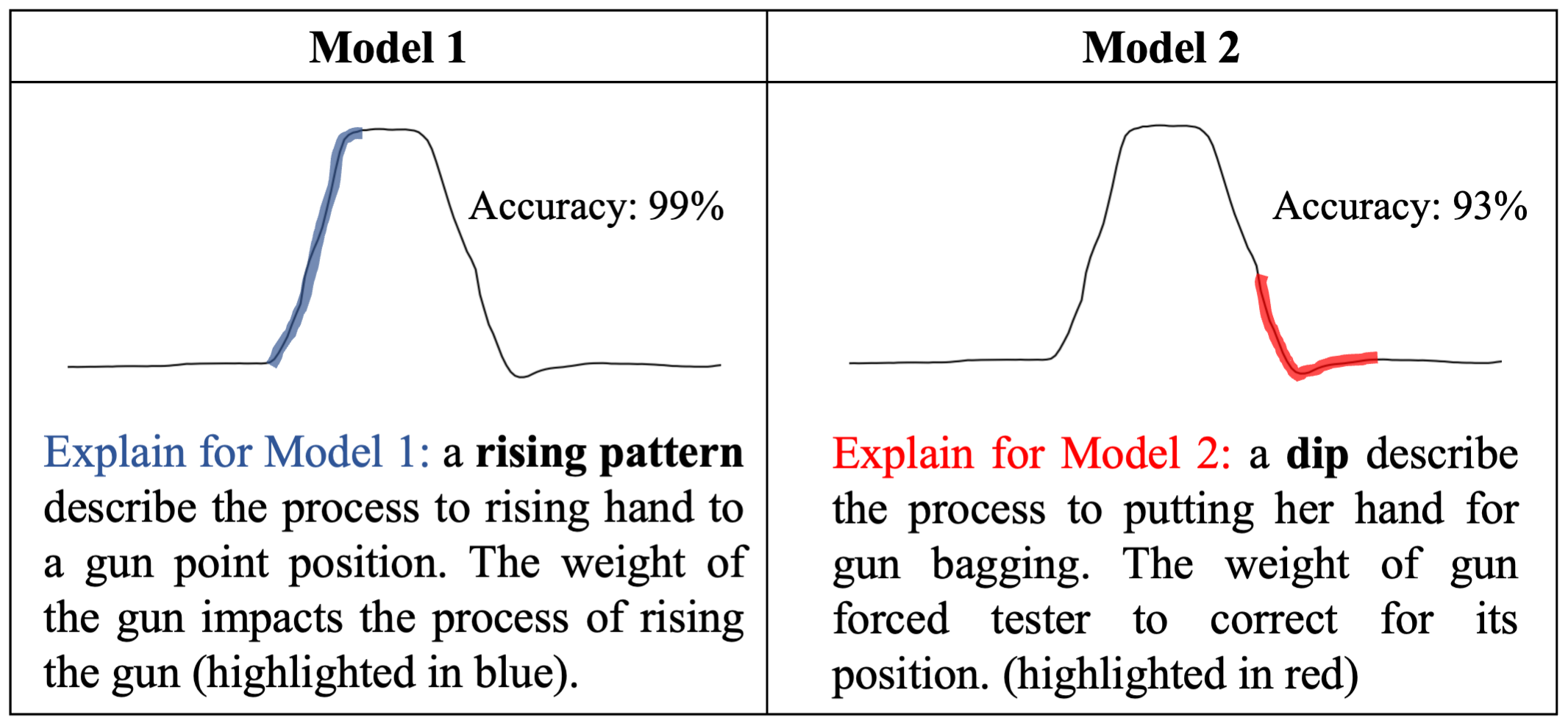}
    \caption{Model 1 has higher accuracy but depending on the operator's height. Model 2 has lower accuracy but show coherency with the true mechanism. }
\end{figure}

Consider the classical GunPoint classification task (classifying a person holding gun (Gun) vs no gun (No Gun) via the time series sensing the hand position)~\cite{ye2009time}. Suppose through accuracy metric, we identify that two models, Model 1 and Model 2,  obtains accuracy of 99\% and 93\% respectively in the evaluation test. Solely based on this evaluation, users might consider Model 1 is better than Model 2 (Fig.1 top). However, if we reveal the logic and features that the decision making of the model to the user as illustrated in Fig. 1.bottom, some previously overlooked concerns may be raised, and potentially changing the evaluation result. Will the first model actual partially make the decision based on height or arm length? As shown in the figure, Model 1 might not be able to generalize to a person with a different height or arm length. We also will re-evaluate the performance of Model 2 --- the `dip' pattern, which is caused by the weight of the gun, can potentially be a better way to identify guns since it has less correlation between the bio-information of a person. This example show that a logically appealed explanation can significantly uncover previously unknown issues which cannot be reflected by metrics such as accuracy. 

It is worth noting that the need of explanation for time series data mining should distinguish from discovering/integrating underlying physical system~\cite{pombo2022benchmarking} or model driven explanation~\cite{selvaraju2016grad,ribeiro2016should}. For example, in Gunpoint data, we are not interested in physiologically why a tester would hold a gun, nor the physical dynamic behind the airflow (e.g. physical system). Instead, time series data mining needs to seek \textit{evidence-based explanations} -- detect the `what' as the sensor data on the little `dip' as evidence, and verify with our existing knowledge: without gun, the hand could `overshoot' and with gun, the data should be relatively smooth. This fact is invariant to the individual's other information such as weight and height. This fact can be called `knowledge' (different from solely model-based explanation~\cite{selvaraju2016grad,ribeiro2016should}, the definition of knowledge not only relies on model, but also relies on why the data formed). From the same example, one can peek the difference between the physical-oriented model and our data mining model, which is evidence-seeking. 

\section{Why should the community ponder over it? Why now?}

Recently, a considerable amount of attention has been given to developing time series foundational model. Such models require strong generalization and the ability to adapt to various downstream tasks. The proposed blue sky idea aims to tackle the key challenge in developing such model --- how to evaluate the generalization performance of the model in time series data. A proper evaluation on time series model will have significant impacts on such foundational model design and have the potential to guide the community toward developing time series models that foster convergent research.

\noindent \textbf{To what degree does the detected explanation represent knowledge? }
To what degree it is considered our finding as verified knowledge? There is often disagreement about whether the knowledge discovered by machine learning models counts towards knowledge. For example, a physicist may hope to establish an understanding on the mechanisms using
their own knowledge and `intelligence' to inform
their models~\cite{carleo2019machine}, instead of getting the intelligence extracting it from data. Ideally, it should help ground the time series with the application. The goal of knowledge and explanations from data mining models is to assist domain researchers without compromising conventional standards \cite{doshi2017towards}.

\noindent\textbf{How to ensure the quality of explanation?} \textit{Explanations are not equal.} While the structure of explanation are seeking for general patterns and domain knowledge driven, instead of restricted causal learning~\cite{lombrozo2006structure, keil2006explanation}, how can we ensure the quality of explanations are with \textit{simple, exact, fruitful, and efficient} explanations~\cite{salmon2006four}, so they could achieve the desirable satisfaction~\cite{ahn2003understanding} and foster mutual advance~\cite{hickling2001emergence}? 

\noindent\textbf{How can we make cheap and scalable knowledge-coherent model explanations? } Time series data mining has a long history of `case study' based evaluation \cite{keogh2005hot,lin2003clustering,rakthanmanon2012mdl,zhang2022joint,zhang2022robust} --- visually explanation of evidence of findings in the real world application, and considered as the best way to share knowledge with domain expertise. However, human evaluation is hard to perform in scale, especially under the fast-growing number of new models, and diverse mechanisms in different domains. This problem is even more severe in time series due to expertise scarcity given the cost of obtaining knowledge is expensive (requiring years of training in a specific field).

\section{What will success look like?} 
Given the inherent challenges of time series data, the success of this project hinges on developing a human-in-the-loop, knowledge-centric evaluation protocol tailored for time series data mining. This protocol will enable an accurate assessment of time series model generalization, reducing unnecessary development costs caused by flawed evaluations and incomplete knowledge. It will encourage data holders to share not only data but also domain knowledge and verification mechanisms for research purposes. Furthermore, the protocol will facilitate precise comparison and selection of time series tools, fostering collaboration between AI researchers and domain experts to refine solutions and drive scientific discovery based on time series.
\section*{Acknowledgment}
The author would like to thank the Blue Sky reviewers for taking the time to read the script. This work has been supported in part by NSF awards CNS-2431514. 

\bibliography{explainable}

\begin{thebibliography}{10}

\bibitem{ahn2003understanding}
W.-k. Ahn, L.~R. Novick, and N.~S. Kim.
\newblock Understanding behavior makes it more normal.
\newblock {\em Psychonomic Bulletin \& Review}, 10(3):746--752, 2003.

\bibitem{carleo2019machine}
G.~Carleo, I.~Cirac, K.~Cranmer, L.~Daudet, M.~Schuld, N.~Tishby, L.~Vogt-Maranto, and L.~Zdeborov{\'a}.
\newblock Machine learning and the physical sciences.
\newblock {\em Reviews of Modern Physics}, 91(4):045002, 2019.

\bibitem{chen2020simple}
T.~Chen, S.~Kornblith, M.~Norouzi, and G.~Hinton.
\newblock A simple framework for contrastive learning of visual representations.
\newblock In {\em International conference on machine learning}, pages 1597--1607. PMLR, 2020.

\bibitem{deng2009imagenet}
J.~Deng, W.~Dong, R.~Socher, L.-J. Li, K.~Li, and L.~Fei-Fei.
\newblock Imagenet: A large-scale hierarchical image database.
\newblock In {\em 2009 IEEE conference on computer vision and pattern recognition}, pages 248--255. Ieee, 2009.

\bibitem{dogan2021machine}
A.~Dogan and D.~Birant.
\newblock Machine learning and data mining in manufacturing.
\newblock {\em Expert Systems with Applications}, 166:114060, 2021.

\bibitem{doshi2017towards}
F.~Doshi-Velez and B.~Kim.
\newblock Towards a rigorous science of interpretable machine learning.
\newblock {\em arXiv preprint arXiv:1702.08608}, 2017.

\bibitem{esling2012time}
P.~Esling and C.~Agon.
\newblock Time-series data mining.
\newblock {\em ACM Computing Surveys (CSUR)}, 45(1):1--34, 2012.

\bibitem{hempel1948studies}
C.~G. Hempel and P.~Oppenheim.
\newblock Studies in the logic of explanation.
\newblock {\em Philosophy of science}, 15(2):135--175, 1948.

\bibitem{hickling2001emergence}
A.~K. Hickling and H.~M. Wellman.
\newblock The emergence of children's causal explanations and theories: evidence from everyday conversation.
\newblock {\em Developmental psychology}, 37(5):668, 2001.

\bibitem{inman2013solar}
R.~H. Inman, H.~T. Pedro, and C.~F. Coimbra.
\newblock Solar forecasting methods for renewable energy integration.
\newblock {\em Progress in energy and combustion science}, 39(6):535--576, 2013.

\bibitem{keil2006explanation}
F.~C. Keil.
\newblock Explanation and understanding.
\newblock {\em Annu. Rev. Psychol.}, 57(1):227--254, 2006.

\bibitem{keogh2005hot}
E.~Keogh, J.~Lin, and A.~Fu.
\newblock Hot sax: Efficiently finding the most unusual time series subsequence.
\newblock In {\em Fifth IEEE International Conference on Data Mining (ICDM'05)}, pages 8--pp. Ieee, 2005.

\bibitem{lin2003clustering}
J.~Lin, E.~Keogh, and W.~Truppel.
\newblock Clustering of streaming time series is meaningless.
\newblock In {\em Proceedings of the 8th ACM SIGMOD workshop on Research issues in data mining and knowledge discovery}, pages 56--65, 2003.

\bibitem{lombrozo2006structure}
T.~Lombrozo.
\newblock The structure and function of explanations.
\newblock {\em Trends in cognitive sciences}, 10(10):464--470, 2006.

\bibitem{muller2024data}
M.~Muller.
\newblock Data silences: How to unsilence the uncertainties in data science.
\newblock In {\em Proceedings of the 2024 SIAM International Conference on Data Mining (SDM)}, pages 388--391. SIAM, 2024.

\bibitem{pombo2022benchmarking}
D.~V. Pombo, P.~Bacher, C.~Ziras, H.~W. Bindner, S.~V. Spataru, and P.~E. S{\o}rensen.
\newblock Benchmarking physics-informed machine learning-based short term pv-power forecasting tools.
\newblock {\em Energy Reports}, 8:6512--6520, 2022.

\bibitem{queen2024encoding}
O.~Queen, T.~Hartvigsen, T.~Koker, H.~He, T.~Tsiligkaridis, and M.~Zitnik.
\newblock Encoding time-series explanations through self-supervised model behavior consistency.
\newblock {\em Advances in Neural Information Processing Systems}, 36, 2024.

\bibitem{rakthanmanon2012mdl}
T.~Rakthanmanon, E.~J. Keogh, S.~Lonardi, and S.~Evans.
\newblock Mdl-based time series clustering.
\newblock {\em Knowledge and information systems}, 33:371--399, 2012.

\bibitem{ribeiro2016should}
M.~T. Ribeiro, S.~Singh, and C.~Guestrin.
\newblock " why should i trust you?" explaining the predictions of any classifier.
\newblock In {\em Proceedings of the 22nd ACM SIGKDD international conference on knowledge discovery and data mining}, pages 1135--1144, 2016.

\bibitem{salmon2006four}
W.~C. Salmon.
\newblock {\em Four decades of scientific explanation}.
\newblock University of Pittsburgh press, 2006.

\bibitem{selvaraju2016grad}
R.~R. Selvaraju, A.~Das, R.~Vedantam, M.~Cogswell, D.~Parikh, and D.~Batra.
\newblock Grad-cam: Why did you say that?
\newblock {\em arXiv preprint arXiv:1611.07450}, 2016.

\bibitem{shokoohi2015discovery}
M.~Shokoohi-Yekta, Y.~Chen, B.~Campana, B.~Hu, J.~Zakaria, and E.~Keogh.
\newblock Discovery of meaningful rules in time series.
\newblock In {\em Proceedings of the 21th ACM SIGKDD international conference on knowledge discovery and data mining}, pages 1085--1094, 2015.

\bibitem{siddiquee2019seismo}
M.~A. Siddiquee, Z.~Akhavan, and A.~Mueen.
\newblock Seismo: Semi-supervised time series motif discovery for seismic signal detection.
\newblock In {\em Proceedings of the 28th ACM International Conference on Information and Knowledge Management}, pages 99--108, 2019.

\bibitem{sun2020review}
C.~Sun, S.~Hong, M.~Song, and H.~Li.
\newblock A review of deep learning methods for irregularly sampled medical time series data.
\newblock {\em arXiv preprint arXiv:2010.12493}, 2020.

\bibitem{wang2019review}
H.~Wang, Z.~Lei, X.~Zhang, B.~Zhou, and J.~Peng.
\newblock A review of deep learning for renewable energy forecasting.
\newblock {\em Energy Conversion and Management}, 198:111799, 2019.

\bibitem{wang2017earthquake}
Q.~Wang, Y.~Guo, L.~Yu, and P.~Li.
\newblock Earthquake prediction based on spatio-temporal data mining: an lstm network approach.
\newblock {\em IEEE Transactions on Emerging Topics in Computing}, 8(1):148--158, 2017.

\bibitem{wu2021current}
R.~Wu and E.~J. Keogh.
\newblock Current time series anomaly detection benchmarks are flawed and are creating the illusion of progress.
\newblock {\em IEEE transactions on knowledge and data engineering}, 35(3):2421--2429, 2021.

\bibitem{ye2009time}
L.~Ye and E.~Keogh.
\newblock Time series shapelets: a new primitive for data mining.
\newblock In {\em Proceedings of the 15th ACM SIGKDD international conference on Knowledge discovery and data mining}, pages 947--956, 2009.

\bibitem{yue2022ts2vec}
Z.~Yue, Y.~Wang, J.~Duan, T.~Yang, C.~Huang, Y.~Tong, and B.~Xu.
\newblock Ts2vec: Towards universal representation of time series.
\newblock In {\em Proceedings of the AAAI Conference on Artificial Intelligence}, volume~36, pages 8980--8987, 2022.

\bibitem{zeng2023transformers}
A.~Zeng, M.~Chen, L.~Zhang, and Q.~Xu.
\newblock Are transformers effective for time series forecasting?
\newblock In {\em Proceedings of the AAAI conference on artificial intelligence}, volume~37, pages 11121--11128, 2023.

\bibitem{zhang2022joint}
L.~Zhang, N.~Patel, X.~Li, and J.~Lin.
\newblock Joint time series chain: Detecting unusual evolving trend across time series.
\newblock In {\em Proceedings of the 2022 SIAM International Conference on Data Mining (SDM)}, pages 208--216. SIAM, 2022.

\bibitem{zhang2022robust}
L.~Zhang, Y.~Zhu, Y.~Gao, and J.~Lin.
\newblock Robust time series chain discovery with incremental nearest neighbors.
\newblock In {\em 2022 IEEE International Conference on Data Mining (ICDM)}, pages 1311--1316. IEEE, 2022.

\bibitem{zhang2022self}
X.~Zhang, Z.~Zhao, T.~Tsiligkaridis, and M.~Zitnik.
\newblock Self-supervised contrastive pre-training for time series via time-frequency consistency.
\newblock {\em Advances in Neural Information Processing Systems}, 35:3988--4003, 2022.

\bibitem{zhu2019introducing}
Y.~Zhu, M.~Imamura, D.~Nikovski, and E.~Keogh.
\newblock Introducing time series chains: a new primitive for time series data mining.
\newblock {\em Knowledge and Information Systems}, 60:1135--1161, 2019.

\bibitem{zhu2016matrix}
Y.~Zhu, Z.~Zimmerman, N.~S. Senobari, C.-C.~M. Yeh, G.~Funning, A.~Mueen, P.~Brisk, and E.~Keogh.
\newblock Matrix profile ii: Exploiting a novel algorithm and gpus to break the one hundred million barrier for time series motifs and joins.
\newblock In {\em 2016 IEEE 16th international conference on data mining (ICDM)}, pages 739--748. IEEE, 2016.

\end{thebibliography}
\bibliographystyle{abbrv}

\end{document}